\documentclass[a4paper, 10pt, conference]{ieeeconf}      

\IEEEoverridecommandlockouts                              
\usepackage{graphics} 
\usepackage{epsfig} 
\usepackage{mathptmx} 
\usepackage{times} 
\usepackage{amsmath} 
\usepackage{amssymb}  
\usepackage{url}
\usepackage{float}
\usepackage{color}
\usepackage[pdftex,
  bookmarks=false,
  bookmarksnumbered=true,
  bookmarksopen=true,
  hypertexnames=true,
  breaklinks=true,
  colorlinks=true,
  linkcolor=red,
  citecolor=green,
  pagecolor=blue,
  pagebackref=true,
  urlcolor=red]{hyperref}

\title{\Large\bf
A Deep Network for Arousal-Valence Emotion Prediction with Acoustic-Visual Cues 
}

\author{Songyou Peng$^{1}$ \qquad Le Zhang$^{1}$ \qquad Yutong Ban$^{2}$ \qquad Meng Fang$^{3}$ \qquad Stefan Winkler$^{1}$
\\
{$^{1}$ADSC-UIUC\qquad $^{2}$INRIA Grenoble\qquad $^{3}$Tencent AI Lab}\\
{\tt\footnotesize \{songyou.peng, zhang.le, stefan.winkler\}@adsc-create.edu.sg 
}\\
{\tt\footnotesize yutong.ban@inria.fr, moefang@gmail.com}
}

\begin{document}

\maketitle
\thispagestyle{empty}
\pagestyle{empty}
\section{Introduction}
In this paper, we comprehensively describe the methodology of our submissions to the One-Minute Gradual-Emotion Behavior Challenge (OMG-Emotion).
Section~\ref{sec:data} introduces the representation of videos and audios that we use as the input of deep networks. 
The designation of model architectures are depicted in section~\ref{sec:data}, followed by the results in section~\ref{sec:result} and conclusion in Section~\ref{sec:con}. Source codes for this paper are available~\footnote{\url{https://github.com/pengsongyou/OMG-ADSC}}.

\section{Data Representation}
\label{sec:data}
In our two submissions, our models use either only visual input or both visual and acoustic input.
This section details how we preprocess these two modalities from the provided OMG-emotion dataset~\cite{barros2018omg}.

\subsection{Acoustic Representation}
Since the audio files are not provided separately, we first convert all snippets to WAV files, each one of which is single-channel and sampled at 16kHz.
Similar to~\cite{Nagrani17}, spectrograms are then calculated every 10ms with a sliding hamming window of width 25ms and 512-point FFT.
We assume that 3 seconds of the audio signal contains emotion information, thus 3 seconds of audio signal is taken to get the short-time Fourier transform (STFT) spectrum. Since each frequency bin in the spectrum is a complex number, we obtain an STFT map of size $257\times 300 \times 2$, where $2$ indicates both the real and imaginary parts of the acquired STFT values. 

\subsection{Visual Representation}
For each utterance in the dataset, we initially extract all frames with OpenCV~\cite{opencv_library}. \emph{MTCNN}~\cite{zhang2016joint} is then applied to detect and align faces.
Since we employ \emph{SphereFace}~\cite{liu2017sphereface} as the backbone network for videos, all face images are resized to $112\times 96\times 3$.

\section{Network Architecture}
\label{sec:net}
\begin{figure*}
    \centering
    \includegraphics[width=0.92\textwidth]{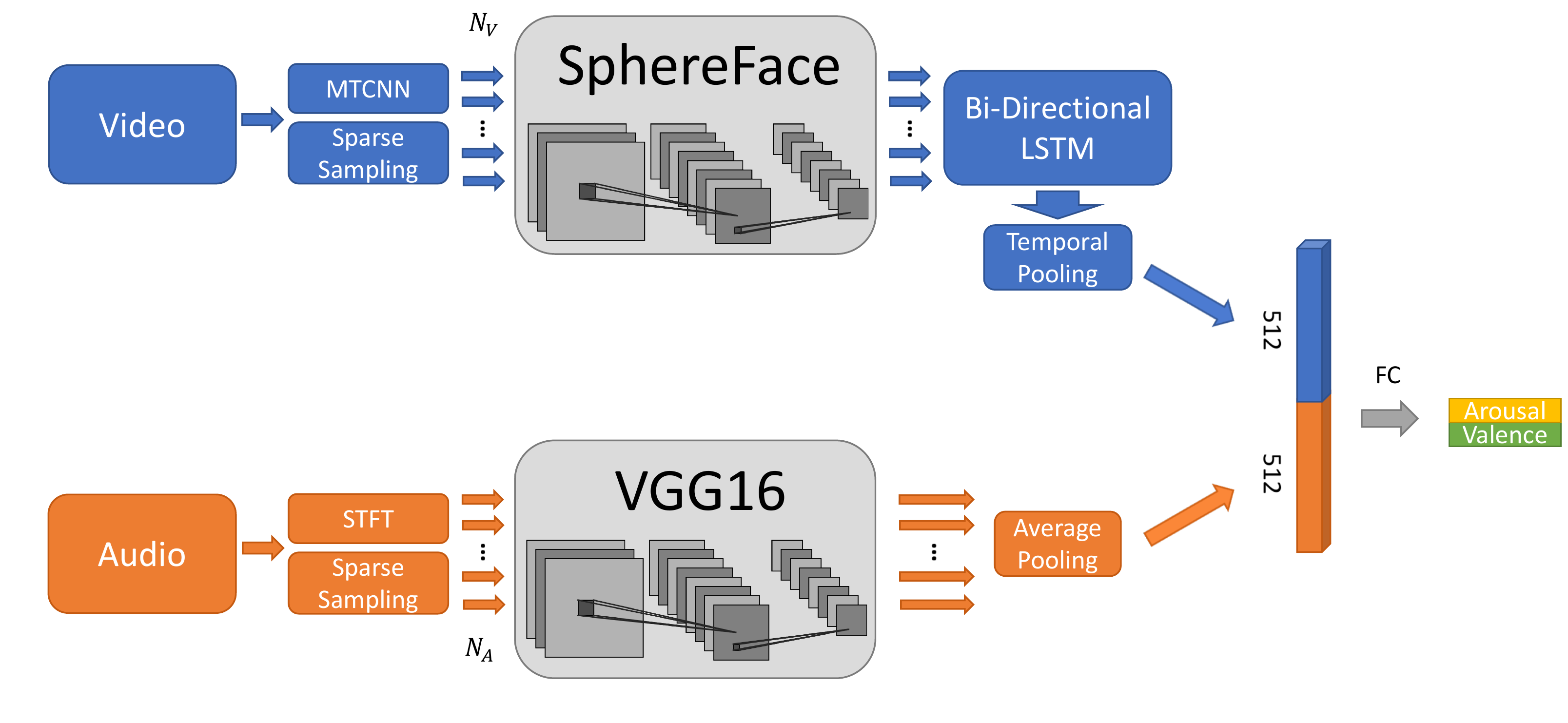}
    \caption{The workflow of the joint training architecture.}
    \label{fig:workflow}
\end{figure*}

With the preprocessed video and audio data, we design and adopt the following deep networks to the arousal-valence regression problem.
The overall architecture can be viewed in figure~\ref{fig:workflow}.

\subsection{Audio Model (ANet)}
We use a rather straightforward network for the audio stream.
The STFT maps is input to the base network \emph{VGG-16}~\cite{vgg16} pretrained on ImageNet. Since the depth dimension of an STFT map is $2$, we modified the first layer of \emph{VGG-16} accordingly.
The $7\times 9\times 512$ output feature is then fed into two fully-connected (FC) layers with dropouts in between. 

If we solely train the network on audios, another FC layer and the Tanh function are applied to acquire the final arousal and valence values.

\subsection{Video Model (VNet)}
Due to the fact that the length of every snippet in dataset varies from $1-2$ seconds up to $20-30$ seconds, the number of extracted frames may be quite different among snippets.
In order to fully utilize the temporal information with feasible GPU memory utilization, we first sparsely sample $N_V$ frames from a snippet.
Inspired by~\cite{tsn2016eccv}, we divide a snippet in order into $N_V$ segments, from which a single frame is randomly sampled. 

For each of the selected frames, an intermediate feature (dim=512) can be obtained from \emph{SphereFace}. Then, these $N_V$ features are fed into a bidirectional LSTM. Finally, an temporal-average pooling layer followed by an FC and Tanh are employed to acquire two emotion scores.

\subsection{Audio-Video Joint Training}
For the sake of jointly training ANet and VNet together, we design the following scheme.
First we take the VNet and ANet solely trained beforehand.
With VNet kept the same, we require to sample $N_A$ STFT maps from every snippet and then average their outputs of the penultimate FC layer in ANet.
Then, we simply concatenate the ANet and VNet features and feed into another FC layer followed by Tanh.
Figure~\ref{fig:workflow} illustrates the architecture of the joint training.

\subsection{Implementation details}
The architectures are implemented in PyTorch~\cite{paszke2017automatic}. 
In joint training, We train with an initial learning rate of $0.001$ and decrease by a factor of 10 every 7 epochs. For each mini-batch, $N_A$ and $N_V$ are set to 4 and 16, respectively. Batch size is 6.
We also sue gradient clipping when the norm is over 20.
With one NVIDIA GTX TITAN X, it takes around 7 minutes for one epoch.

It should be noted that one important difference between joint training and training VNet and ANet separately is the loss function. MSE loss is employed for sole training while CCC for joint training.

\section{Results}
\label{sec:result}
In the part, we briefly illustrate the performance of our models over the provided baseline methods in CCC.

Table~\ref{tab:audio} compares our ANet with a method~\cite{barros2016developing} pretrained on RAVDESS and another method on OpenSmile. It shows that even without pretraining on any audio dataset, our ANet still outperforms the baselines in both arousal and valence scores.

\begin{table}[!ht]
\centering
\caption{CCC Comparison when using only Audio. ``OS'' indicates using OpenSmile Dataset}
\label{tab:audio}
\begin{tabular}{l|l|l|l}
\hline
                           & Arousal & Valence & Total  \\ \hline\hline
Baseline~\cite{barros2016developing}        & 0.08    & 0.10    & 0.18   \\ \hline
Baseline (OS) & 0.15    & 0.21    & 0.36   \\ \hline
\textbf{Our ANet}   & \textbf{0.1879}  & \textbf{0.256}   & \textbf{0.4439} \\ \hline
\end{tabular}
\end{table}

In table~\ref{tab:video}, VNet shows superior performance over the baseline~\cite{barros2016developing}. VNet doubles both CCC of arousal and valence.

\begin{table}[!ht]
\centering
\caption{CCC Comparison when using only Video (submission \#1)}
\label{tab:video}
\begin{tabular}{l|l|l|l}
\hline
                           & Arousal & Valence & Total  \\ \hline\hline
Video Baseline~\cite{barros2016developing}             & 0.12    & 0.23    & 0.35   \\ \hline
\textbf{Our VNet}      & \textbf{0.2798}  & \textbf{0.4688}  & \textbf{0.7486} \\ \hline
\end{tabular}
\end{table}

Finally, table~\ref{tab:joint} illustrates the effectiveness of jointly training networks for both audio and video streams. Further performance improvement has been achieved with such a training scheme.
\begin{table}[!ht]
\centering
\caption{CCC values of our Audio-Video joint training (submission \#2)}
\label{tab:joint}
\begin{tabular}{l|l|l|l}
\hline
                           & Arousal & Valence & Total  \\ \hline\hline
\textbf{Joint Training}    & \textbf{0.3036}  & \textbf{0.4796}  & \textbf{0.7832} \\ \hline
\end{tabular}
\end{table}

\section{Conclusion}
\label{sec:con}
This paper describes a novel architecture for arousal-valence estimation on the OMG-Emotion Dataset.
We have shown the advantage of our deep network over the baseline methods.

\section*{Acknowledgment}
This research is supported by the SERC Strategic Fund from the
Science and Engineering Research Council (SERC), A*STAR (project
no. a1718g0046).
\bibliographystyle{plain}
\bibliography{ref}

\end{document}